\documentclass[11pt,letterpaper,copyright]{perplexity}

\usepackage[numbers,sort&compress]{natbib}
\usepackage{graphicx}
\usepackage{xurl}
\usepackage{hyperref}

\usepackage{graphicx}
\usepackage{xcolor}
\usepackage{changepage}
\usepackage{framed}

\definecolor{pplxInky}{HTML}{133B39}   %
\definecolor{pplxPeacock}{HTML}{2E5E5A}%
\definecolor{pplxTurq}{HTML}{20808D}   %
\definecolor{pplxPlex}{HTML}{2EE5EA}   %

\definecolor{pplxBg}{HTML}{F3F8F8}

\newenvironment{kkboxline}{%
  \MakeFramed{%
    \setlength{\fboxsep}{4pt}%
    \advance\hsize-\width
    \FrameRestore
  }%
  \noindent\ignorespaces
}{%
  \endMakeFramed
}

\usepackage{cleveref}
\crefname{section}{Section}{Sections}
\Crefname{section}{Section}{Sections}
\hypersetup{
  linkcolor=black,
  citecolor=blue,
  urlcolor=blue
}

\uselogo{}

\makeatletter
\let\theabstract\@empty
\makeatother

\newcommand{\mypara}[1]{\medskip\noindent\textbf{#1.}}
\newcommand{\eg}{\emph{e.g.}}

\makeatletter
\renewcommand\normalsize{%
  \@setfontsize\normalsize{10.5pt}{12.6pt}%
  \abovedisplayskip 10.5pt plus 2pt minus 5pt
  \abovedisplayshortskip \z@ plus 3pt
  \belowdisplayshortskip 6pt plus 3pt minus 3pt
  \belowdisplayskip \abovedisplayskip
  \let\@listi\@listI}
\makeatother

\AtBeginDocument{%
  \normalsize
  \setlength{\parskip}{\baselineskip}%
}

\renewcommand\Affilfont{\small\centering}

\title{\centering
  \textbf{Security Considerations for Artificial Intelligence Agents} \\
  (Perplexity Response to NIST/CAISI Request for Information 2025-0035)
}
\date{March 9, 2026}

\author[{$\dagger\S$}]{Ninghui Li}
\author[{$\dagger\S$}]{Kaiyuan Zhang}
\author[{$\dagger$}]{Kyle Polley}
\author[{$\dagger$}]{Jerry Ma}

\affil[$\dagger$]{Perplexity}
\affil[$\S$]{Purdue University}

\begin{document}

\sloppypar
\maketitle

\section*{Abstract}
This article, a lightly adapted version of Perplexity's \href{https://www.regulations.gov/comment/NIST-2025-0035-0505}{response} to NIST/CAISI Request for Information 2025-0035~\cite{nist_rfi_ai_agents}, details our observations and recommendations concerning the security of frontier AI agents. These insights are informed by Perplexity's experience operating general-purpose agentic systems used by millions of users and thousands of enterprises in both controlled and open-world environments~\cite{perplexity_comet,perplexity_computer}. Agent architectures change core assumptions around code-data separation, authority boundaries, and execution predictability, creating new confidentiality, integrity, and availability failure modes. We map principal attack surfaces across tools, connectors, hosting boundaries, and multi-agent coordination, with particular emphasis on indirect prompt injection, confused-deputy behavior, and cascading failures in long-running workflows. We then assess current defenses as a layered stack: input-level and model-level mitigations, sandboxed execution, and deterministic policy enforcement for high-consequence actions. Finally, we identify standards and research gaps, including adaptive security benchmarks, policy models for delegation and privilege control, and guidance for secure multi-agent system design aligned with NIST risk management principles.

\section{Security Threats, Risks, and Vulnerabilities Affecting AI Agent Systems}\label{sec:1}

\subsection{Security Considerations Posed by AI Agent Systems}

\begin{kkboxline}
\textbf{1(a):} 
What are the unique security threats, risks, or vulnerabilities currently affecting AI agent systems, distinct from those affecting traditional software systems?
\end{kkboxline}

\mypara{First, AI agent systems further blur the line between code and data} The separation of code and data is a fundamental principle in computer security. It means that data generally should not be treated as executable code, and code should not be dynamically altered by user-controlled input. Enforcing this boundary is important for securing modern software systems.

The von Neumann architecture, which underlies today's digital computers, introduced the stored-program concept: Instructions (code) and data are stored in the same memory and share the same pathways. This is the root cause of many software security vulnerabilities such as buffer overflow and code injection attacks. In desktop computing platforms such as Windows, macOS, and Linux, code (in the form of application programs) is trusted, and security mechanisms focus on protecting the integrity of the platforms against potentially malicious data, typically coming from the internet. Similarly, in web applications, enforcing strong separation of code and data is the linchpin of defense against SQL injection, Cross-Site Scripting, Cross-Site Request Forgery, and other injection attacks. In mobile computing platforms such as Android and iOS, code (in the form of apps) developed by potentially unknown developers needs to be added to the system. As a result, security mechanisms include using sandboxed environments to run apps, preventing access to low-level operating system resources as well as controlling their access to application-level resources via permissions.

Each generation of computing platforms introduces new code-data separation problems; Large Language Model (LLM)-powered agent systems represent the latest and perhaps most severe instance of this recurring challenge. 
In LLM-powered AI agent systems, the distinction between code and data is further blurred. Plaintext prompts play the role of code, shaping LLMs' control flows by, \eg{}, steering the invocation of tool calls. 
In addition, dynamically generated text can itself become a prompt for an LLM, so the control flow depends on payloads that are unknown until runtime.
One can view LLMs as providing a new layer of programmable computing interfaces, and there is no binary distinction between code and data at this layer. For example, Agent Skills can be viewed as code libraries for this new programming interface provided by LLMs~\cite{agentSkillsSpec2025,xu2026agentSkills}.

\mypara{Second, AI agent systems have significantly more flexible automation} Traditional software systems mostly follow pre-programmed workflows, and the resulting automation is narrow, predictable, and explicitly specified by developers. On the other hand, AI agents (especially LLM-based or planning-based agents) operate differently. They can accept high-level goals and decide which intermediate steps to take. Rather than executing a pre-programmed workflow, they can dynamically construct workflows, choose which APIs/tools to call, and chain actions based on intermediate results. They can generalize to unseen but related situations, interpret ambiguous input, and handle partial information.

The consequences are manifold. For AI agents to be useful, they must often be granted broad capabilities---such as accessing file systems, querying databases, using API credentials, executing code, and conducting transactions. Misuse of these capabilities, whether malicious or accidental, opens the door to significant vulnerabilities.
A concrete example is OpenClaw~\cite{openclaw_doc}, 
an open-source AI assistant that connects AI models to local files and messaging platforms like WhatsApp and Discord to automate tasks continuously. Several security incidents involving OpenClaw have been documented, including attacks recorded as CVE-2026-25253 and CVE-2026-26327~\cite{CVE-2026-25253,CVE-2026-26327}. Compared to conventional software, logic driven by LLMs is inherently non-deterministic and more opaque. As a result, it becomes more difficult to reason about reachable states, enumerate undesirable behaviors, or formally verify system safety.

\mypara{Third, existing security mechanisms are a mismatch for agent-based systems}
In computer security, there are no perfect solutions, only tradeoffs. Because absolute security is unattainable, security mechanisms must be designed for the specific computing environments in which they operate, balancing usability, functionality, and risk. Many existing security mechanisms were developed for pre-agent computing environments with tightly scoped and largely deterministic software behavior. As a result, these mechanisms are not always well suited to agent-based systems, whose autonomy, adaptability, and broader operational scope introduce new security challenges.

In desktop systems, security mechanisms are primarily designed around human users. Human users are typically granted broad access privileges that are sufficient, in principle, to cause significant mischief. This design relies on several assumptions: most users act in good faith; many malicious actors are deterred by auditing and real-world consequences; and humans take actions (whether correct or erroneous) relatively slowly. The residual risks that arise when these assumptions fail have generally been considered tolerable. But agents, which can take actions and exercise privileges at machine speed, may require new security mechanisms to sufficiently mitigate risk.

Similarly, web browsers rely on security mechanisms such as sandboxing and the Same-Origin Policy to isolate different websites. When cross-site interactions do occur, they typically happen through human actions. The underlying assumption is that users can assess the risks of these interactions and act accordingly. With the introduction of agentic browsers, AI agents now routinely perform a growing share of those actions through automated means. Therefore, while some traditional web security mechanisms will translate directly to agentic browsing, the security community will also need to craft novel defenses tailored to the way agents perform actions in web environments.

\mypara{The preceding analysis highlights two key design imperatives for securing AI agent systems}  First, new security abstractions are needed to address the new computational layer introduced by automated agents powered by probabilistic AI models. Such agents act on behalf of human users, yet have significantly different characteristics.
Second, although containment and isolation remain important for securing agent-based systems, traditional principles, such as least privilege and fine-grained access control, must be reexamined and adapted to the dynamic and evolving behavior of agents. We return to these issues later in this response.

\subsection{Security Threats and Risks in AI Agent Systems}

\begin{kkboxline}
\textbf{1(e):}
What unique security threats, risks, or vulnerabilities currently affect multi-agent systems, distinct from those affecting singular AI agent systems?
\end{kkboxline}

We analyze the security threats and risks in AI agent systems from two perspectives. First, we examine the potential security and privacy consequences using the CIA triad: Confidentiality, Integrity, and Availability. We then analyze the possible attack surfaces and sources of system failure that could give rise to these consequences.

\mypara{Confidentiality}  
Confidentiality violations can occur when AI agents access sensitive information and subsequently leak it through channels accessible to unauthorized entities. To enable meaningful automation, agents often require access to a broad range of data, including user credentials for communication, financial, and other services; sensitive personal information such as addresses, phone numbers, email addresses, Social Security numbers, credit card details, medical records, and financial information; as well as business or personal secrets. Such information may reside in locations accessible to agents, including local file systems, cloud storage, email accounts, and communication platforms such as Slack. Within AI agent systems, sensitive data can propagate through multiple internal pathways, including tool outputs, workspace files, memory entries, and webhook responses. Agents may also transfer data across systems through external connectors or browser automation. While these capabilities enable powerful automation, they also create additional opportunities for information leakage or for data to be mistakenly exposed to the wrong user, process, or session.

\mypara{Integrity}
Integrity violations occur when a system's behavior, decisions, data, or actions are altered in unauthorized, unintended, or misleading ways. These violations may involve unwanted modifications to critical systems (\eg{}, deleting or altering files, installing software, or modifying configurations) as well as unintended communications or transactions with external services (\eg{}, sending emails or initiating financial transfers).
Integrity violations can also manifest as the completion of tasks in a substandard or manipulated manner, potentially due to malicious influence. For example, an agent might procure a product or service from one vendor even though another vendor offers the same quality at a lower price.
More broadly, integrity violations may include presenting human users with erroneous or misleading information. Such information can lead users to make incorrect decisions later, even outside the AI agent system itself.

\mypara{Availability}
Availability failures arise when system resources are excessively consumed or when AI agent systems become unable to serve users. In agent-based systems, long-running tasks, scheduled jobs, browser automation, and external connectors introduce additional dependencies that may fail and block workflows. Resource overconsumption can also cause tasks to terminate prematurely, while partial failures may leave an agent in a stalled state or trapped in repeated execution loops.
As agent systems scale to support longer-running, multi-step workflows, their exposure to both accidental cascading failures and deliberate resource-exhaustion attacks increases. In cascading failure scenarios, an orchestrator may decompose a task across multiple sub-agents or tool invocations; the failure of a single component can then propagate through the pipeline, causing downstream agents to stall, generate erroneous outputs, or enter repeated retry loops that place additional load on shared infrastructure.
From an adversarial perspective, denial-of-service attacks~\cite{li2025thinktrap} can also target the inherently high computational cost of LLM inference, exploiting it to exhaust system resources and degrade availability.

\mypara{Attack Surfaces}  Attack surfaces are any inputs or interaction channels that adversaries can control and potentially exploit to cause violations of Confidentiality, Integrity, or Availability. Several classes of adversaries are relevant in AI agent systems:

\begin{itemize}
    \vspace*{-0.08in}
\item \textbf{External content providers.} These include entities that control information in external systems with which agentic systems interact. Examples include operators of web pages and services, as well as parties who can generate email messages, online posts, reviews, or other content that becomes available to agents through external platforms.
    
\item \textbf{Component providers.} These include providers of core components used in agentic systems, such as LLM services, memory services, tools, or reusable skills.

\item \textbf{Network-based adversaries.} Attackers on the network path can target API endpoints, webhooks, or the communication channels between system components.

\item \textbf{Insiders.} Developers or operators of the agentic system may inadvertently introduce vulnerabilities through software bugs, misconfigurations, or misuse of privileged access.

\item \textbf{Client-side adversaries.} These include entities with access to the client devices through which users interact with AI agent systems.
\end{itemize}

A concrete and increasingly common attack vector is \emph{indirect prompt injection}~\cite{greshake2023not,liu2024formalizing}, in which an adversary embeds adversarial instructions within content that an agent retrieves during normal operation---for example, a web page, an email, or a calendar entry. Because LLMs cannot reliably distinguish trusted instructions from untrusted data, such content can manipulate the agent's behavior. For instance, a poisoned web page could silently instruct a browsing agent to read entries from a user's Google Calendar and exfiltrate them to an attacker-controlled server.

\mypara{System Failures}
Confidentiality, integrity, and availability violations can arise even in the absence of a malicious adversary. In AI agent systems, such violations may result from model errors, flawed agent logic or configuration, incorrect tool implementations, or underlying infrastructure failures. Long-running tasks and recovery mechanisms may replay actions or apply outputs to stale system states. Memory updates may store incorrect facts that propagate into later decisions, while file or code execution tools may introduce unintended changes that are difficult to reverse. These risks are likely to grow as AI agent systems become more complex. As additional tools, connectors, and services are integrated, the number of possible action chains increases, as does the speed at which errors can propagate across interconnected components.

Furthermore, some violations of these principles can occur due to mistakes made by users when installing or operating agentic systems, even when AI agents themselves play no role in the failure. CVE-2026-25253 documents an instance of a one-click remote code execution attack on a local agent~\cite{CVE-2026-25253}, in which the attack sequence did not involve any LLM-driven agent behavior. This example illustrates that, when analyzing the security risks of AI agent systems, it is necessary to look beyond the agents themselves and consider broader architectural changes introduced to enable agent capabilities.

\mypara{Multi-Agent Systems}
Multi-agent systems introduce additional attack surfaces and operational risks. In general-purpose AI agent architectures, subagents and background agents often operate with partial context and exchange information through shared workspaces, memory stores, or tool outputs. This design increases the likelihood of implicit delegation, where one agent indirectly triggers actions by another without a clear chain of authorization. As a result, errors or adversarial instructions can propagate across agents and persist in shared state. These dynamics also complicate auditing and recovery, as decision paths may span multiple agents, intermediate artifacts, and retries, rather than following a single, well-defined execution flow.

Beyond these operational concerns, multi-agent architectures are susceptible to classic confused-deputy vulnerabilities~\cite{hardy1988confused,roychowdhury2024confusedpilot}. An outer agent acting on a user's behalf may be manipulated into instructing a more privileged inner agent to perform actions that neither the user nor the outer agent intended. A related risk is privilege escalation through agent chains: a low-privilege agent can induce a higher-privilege peer to execute sensitive operations, effectively bypassing access controls that would apply if a single agent handled the task end-to-end. Because these interactions occur across loosely defined inter-agent trust boundaries, enforcing consistent authorization policies is difficult. When violations occur, attribution is also challenging, as responsibility is distributed across multiple agents rather than traceable to a single component.

\subsection{Agent Architecture, Deployment and Hosting}

\begin{kkboxline}
\textbf{1(b):} 
How do security threats, risks, or vulnerabilities vary by model capability, agent scaffold software, tool use, deployment method (including internal vs. external deployment), hosting context (including components on premises, in the cloud, or at the edge), use case, and otherwise?
\end{kkboxline}

\mypara{Architecture and Tool Design}
Frontier AI companies are turning these modules into product choices. Model inference APIs have begun exposing built-in tools such as web search, file search, and computer use~\cite{perplexity2026_sonar_api,perplexity2026_agent_api,openai2025_new_tools_agents}. Furthermore, browser use, computer use, connector libraries to third-party services, and other utility features are proliferating at both the product and API layers~\cite{perplexity2026_mcp,openai2025_agentkit,perplexity_comet}. Multi-agent designs, such as those used in Perplexity Computer~\cite{perplexity_computer}, represent a recent architectural innovation that magnifies both the capabilities and complexity of agent systems.

\mypara{Attack Surfaces in Architecture} Adversaries may target one or more attack surfaces in agent architectures:
\begin{itemize}
  \item Tool selection logic, as determined by tool schemas, descriptions, and alternative tool-calling interfaces (\eg{}, CLIs), can be adversarially modified to induce incorrect tool choices by agents.
  \item Tool execution boundaries include hosted tools, local runtime tools, and remote tool servers, and each boundary can introduce untrusted inputs or outputs into the agent loop~\cite{perplexity2026_mcp,openai2025_agents_sdk_tools}.
  \item Web-grounded ingestion includes web search, fetch URL or browse tools, and research mode sources, which bring untrusted content into prompts and tool results~\cite{openai2025_new_tools_agents,perplexity2026_tools_overview,perplexity2026_api_overview}.\footnote{Perplexity's BrowseSafe work~\cite{zhang2025browsesafe} studies prompt injection risks in the context of browser agents.}
  \item Multi-agent coordination surfaces include orchestrators, agent-to-agent messages, and shared sessions or workspaces that pass intermediate results between agents~\cite{openai2025_agentkit,kimi2026_agent_swarm,openclaw_doc}.
  \item Skill and plugin supply chains include plugin code and skill packs that extend capabilities and run with agent privileges~\cite{openclaw_doc}.
\end{itemize}

We discuss one open-source agent platform, OpenClaw, as an illustrative example of the above architecture choices. The platform's Gateway component serves the single source of truth for sessions, routing, and channel connections and supports a multi-channel inbox for messaging platforms, while multi-agent routing binds channels, accounts, or peers to isolated agents with separate sessions and workspaces~\cite{openclaw_doc}. Skills can live in per-agent workspaces or shared locations, plugins can ship skills, and plugins run in-process with the gateway, which means they inherit gateway privileges~\cite{openclaw_doc}. Nodes extend the agent with device-side actions, cron jobs persist scheduled tasks, and the gateway serves a web control UI and a webhook endpoint on the same HTTP server~\cite{openclaw_doc}. Different agent platforms will make different architecture choices, with the resulting security properties heavily shaped by those choices.

\mypara{Consequences and Avenues} From the consequences perspective, architecture determines which data and actions are in scope, which shapes confidentiality, integrity, and availability risks when an agent reads private data, changes external systems, or fails in long workflows~\cite{openai2025_new_tools_agents,anthropic2025_computer_use}. From the avenues perspective, architecture determines the inputs and trust boundaries an attacker can reach, including untrusted messages from channels, web content and browser tasks, tool outputs, and plugin code, as well as the trust placed in model and tool providers~\cite{openai2025_new_tools_agents,anthropic2025_computer_use,openclaw_doc}.
Our BrowseSafe~\cite{zhang2025browsesafe} study on browser agents highlights how untrusted web content can become a direct avenue for prompt injection, which strengthens the case for strict separation between web content and action policies.

\mypara{Deployment and Hosting} Threats change with deployment and hosting. A cloud-hosted agent and its remote browser API add a new trust boundary and new network exposure, while an on-premises or edge deployment reduces some network exposure but increases insider and configuration risk. Local-first deployments keep the gateway on loopback by default, use authenticated tunnels such as Tailscale Serve for remote access, and allow nodes to connect over LAN or tailnet, which shifts exposure from the public internet to device and tunnel security~\cite{openclaw_doc}.
The gateway protocol and pairing workflows control which devices and channels can connect, and the web control UI and webhook endpoint are exposed through the gateway HTTP server, which makes HTTP exposure a hosting decision rather than a model decision~\cite{openclaw_doc}. Some agent systems employ a dedicated virtual machine or container with minimal privileges for computer use, which illustrates the need for strong isolation when agents can control a desktop environment~\cite{anthropic2025_computer_use,perplexity_computer}.

\mypara{Attack Surfaces in Deployment and Hosting} The attack surfaces or avenues are any interfaces or network paths that can be controlled by adversaries. There are different types of adversaries in deployment and hosting:
\begin{itemize}
  \item Network-based adversaries can target exposed services, tunnels, and transport channels when gateways are reachable over LAN or tailnet connections~\cite{openclaw_doc}.
  \item Webhook and web UI exposure create internet-facing endpoints on the gateway HTTP server that can be targeted or misconfigured~\cite{openclaw_doc}.
  \item Device and channel adversaries can exploit pairing or device enrollment workflows if access control is weak or approvals are socially engineered~\cite{openclaw_doc}.
  \item Hosted execution environments, such as code execution or container shells, increase risk when the sandbox or isolation is weak or misconfigured~\cite{anthropic2025_code_execution,openai2025_agents_sdk_tools}.
  \item Desktop control environments are attack surfaces when the agent is allowed to operate a computer, which is why a dedicated VM or container is recommended~\cite{anthropic2025_computer_use}.
\end{itemize}

Architecture and hosting decisions define both the impact of failures and the avenues of attack. Tool surfaces, workflow coordination, and web-grounded research expand what agents can reach, while gateway design, pairing, and hosting choices determine which inputs and networks can reach the agent~\cite{perplexity2026_model_council_blog,perplexity2026_research_site,perplexity2026_api_overview}.
These design choices are therefore first-order security factors that should be evaluated alongside model capability and use case.

\section{Security Practices for AI Agent Systems}\label{sec:2}

\begin{kkboxline}
\textbf{2(a):} 
What technical controls, processes, and other practices could ensure or improve the security of AI agent systems in development and deployment? What is the maturity of these methods in research and in practice?
\end{kkboxline}

\mypara{Security Principles and Defense-in-Depth}
While AI agent systems introduce novel challenges and existing security mechanisms often do not map cleanly to these new contexts, the foundational security principles the community has developed over the past half century remain deeply relevant. We emphasize that these well-established principles should continue to guide the design and protection of AI agent systems, albeit with careful adaptation to the unique properties and risks posed by such systems.

Saltzer and Schroeder's landmark 1975 paper, ``The Protection of Information in Computer Systems''~\cite{saltzer1975protection}, articulated eight enduring design principles for secure systems---among them least privilege, complete mediation, and psychological acceptability. Over the ensuing decades, these principles have shaped the foundations of security education, research, and practice.
Additional principles have since emerged, with defense-in-depth being particularly salient for AI agents. Given the non-deterministic reasoning of large language models (LLMs) and the inherent complexity of modern AI architectures, no single security mechanism can offer comprehensive protection standing alone. Instead, agent developers should employ multiple layers of defense that mutually complement and reinforce each other to limit potential damage should any one layer fail. In recent work~\cite{zhang2025llm}, we argued for systematically applying these principles in the context of agentic systems and demonstrated their practical use through AgentSandbox, a framework designed to secure personal assistant agents.

\mypara{Defenses Against Prompt Injection Attacks}
In the remainder of this section, we focus on defenses against one of the most significant emerging threats in AI agent systems: indirect prompt injection attacks.
Early documented cases of prompt injection~\cite{perez2022ignore_previous_prompt} involved straightforward techniques such as embedding explicit override instructions (\eg{}, ``ignore previous directions'') or role-playing commands that manipulated model behavior. As LLMs evolved to process multiple modalities, including text, images, and audio, researchers also identified attacks that embed malicious directives within non-text inputs. These multimodal prompt injections can evade defenses that rely solely on textual analysis.
Below we discuss defenses against indirect prompt injection attacks across several layers.

\mypara{Input-Level Defenses}
The first layer of defense focuses on preventing attacks or mitigating the effects of attacks before malicious inputs are processed by AI models. Several approaches have been proposed: detecting attacks, removing malicious content from prompts once detected, or modifying inputs through techniques like spotlighting~\cite{hines2024spotlighting} and sandwiching to reduce the impact of attacks on the underlying LLMs.
For example, Liu et al.~\cite{liu2024formalizing} evaluate multiple prompt injection detection strategies, including measuring input perplexity, querying the LLM itself, and verifying the validity of responses. Researchers have also trained specialized detectors to identify malicious prompts, and companies such as ProtectAI and Meta have developed similar detection systems. Some methods leverage internal model signals, including attention patterns~\cite{hung2025_attention_tracker} and activation traces~\cite{abdelnabi2025driftcatchingllmtask}. For surveys, see \eg{},~\cite{rababah2024sok_prompt_hacking,geng2026_prompt_injection_survey,maloyan2026promptinjectionattacksagentic}.

Deploying these detection techniques in production systems presents several challenges. A major issue is the impact of false positives due to the base-rate fallacy~\cite{axelsson2000_base_rate_fallacy}. When benign instances vastly outnumber malicious ones, even a low false positive rate results in most detected ``attacks'' being false alarms, as dictated by Bayes' rule. Handling these detections is nontrivial: simply discarding flagged inputs can severely degrade utility. Performance and cost are additional concerns, as some detection techniques require significant computational resources or rely on access to cutting-edge models, making them expensive or slow to deploy.
These challenges are further amplified in agentic systems, which continuously ingest untrusted content from diverse sources such as web pages, emails, tool outputs, and shared workspaces. The high volume of benign inputs exacerbates the base-rate problem, while the multi-step, latency-sensitive nature of agent workflows means that even modest per-query detection overhead accumulates across the execution chain. As a result, heavyweight detection methods are often impractical for real-time deployment.

There are several open research directions. One key direction is to further improve detection accuracy. Another direction is to develop appropriate response mechanisms after an attack is detected, particularly when false positives are frequent. A related challenge is generating clear explanations for why an input is considered suspicious, potentially enabling LLMs to handle such inputs more effectively. Another important avenue is developing techniques to address multimodal prompt injection threats, which involve malicious content spanning text, images, or other data types.

\mypara{Model-Level Defenses} 
Another layer of defense focuses on making the model itself more resistant to attacks, often by attempting to reestablish a separation between code and data. In virtually all frontier models, inputs are typically structured into distinct roles that guide the model's interpretation of each piece of text or data. Common roles include system, user, and assistant (history), among others. The concept of an ``instruction hierarchy'' trains LLMs to treat instructions from different roles with varying priorities, so that, when conflicts arise, higher-priority instructions take precedence~\cite{wallace2024instruction}.

However, LLMs do not possess a native understanding of role structure and instruction hierarchy. The application layer designates roles using special tokens or structured formatting, which are then flattened into a single token sequence~\cite{wallace2024instruction,chen2025struq}. The model is trained to associate these tokens and structures with authority levels, but because all instructions are ultimately combined into one sequence, there exists no deterministic enforcement layer within the model architecture itself that prevents the model from attending to or following lower-priority segments. Role boundaries remain a learned convention rather than a hard security guarantee, leaving them vulnerable to adversarial influence.
Additionally, lower-priority user instructions are often closest to the model's prediction point, and autoregressive LLMs inherently place greater weight on recent tokens. RLHF further reinforces helpfulness, compliance, and adherence to the user's last request. Unless specifically trained otherwise, this combination of recency and compliance bias can override earlier constraints, causing the model to resolve ambiguity in favor of the most recent directive.

A promising approach is to encode role distinctions at the embedding level. Wu et al.~\cite{wu2025instructional} learn separate embeddings for system instructions, user instructions, third-party data, and generated outputs. Inputs are divided into corresponding segments, and tokens in each segment are associated with their respective embeddings, reinforcing the intended authority hierarchy at a level below the flattened token sequence.

Researchers have also evaluated how well frontier models respect instruction hierarchies~\cite{qin2024infobench,zhang2025iheval,geng2025controlillusion,qin2024sysbench_arxiv,zverev2025can}. Common findings include: (1) Linguistic features, such as tone and structure, often have a stronger influence than the instruction hierarchy, which is unsurprising given that LLMs are trained to understand and follow natural language. (2) More recent LLMs generally perform better at following instruction hierarchies. However, most evaluations focus on conflicting instructions where the conflicts are minor from a security perspective, \eg{}, differences in capitalization, output length, or language. The community would greatly benefit from more thorough assessments of instruction following under security-relevant scenarios.
Taken together, these findings indicate that model-level defenses will be an important piece of the defense, especially as new models are better trained to respect the instruction hierarchy.  However, model-level defenses cannot provide reliable protection on their own. This limitation motivates system-level defenses.

\mypara{Execution Monitoring}
The third layer of defense operates at the system and architecture level, aiming to limit the impact of attacks that bypass input- and model-level defenses. 
As the first step of adding system-level defenses to AI agent systems, mechanisms are needed to run AI agent systems in one or more sandboxes.  Accessing resources from the sandboxes as well as interactions between different sandboxes are controlled with appropriate policies.  For example, one can enforce the policy that an agent's control flow, such as its sequence of tool calls, does not depend on untrusted inputs.

The CaMeL framework~\cite{debenedetti2025defeating} illustrates this principle by explicitly separating control flow from data flow. CaMeL employs a privileged LLM (P-LLM) that processes only the trusted user query and generates a plan in pseudo-Python code. A separate quarantined LLM (Q-LLM) handles untrusted external data, such as emails and web content. A capability-based data-flow tracking system ensures that tainted variables cannot influence privileged operations.  Similar ideas have been explored in several other papers~\cite{tsai2025contextual,an2025ipiguard,li2025drift,wu2025isolategpt}.\footnote{In recent work on AgentSandbox~\cite{zhang2025llm}, we demonstrated how the Saltzer-Schroeder security principles, including least privilege and complete mediation, can be systematically applied to sandbox design for personal assistant agents.} Such approaches, however, depend on being able to distinguish benign control flows from those triggered by attacks and require predefining the set of legitimate flows. In open-ended agent systems, where users can issue novel and complex requests, fully specifying all legitimate control flows remains an open challenge.

\mypara{Deterministic Last Line of Defense}  We argue that it is also necessary to have a deterministic layer that provides hard protection boundaries around components of agent systems.

Unlike input-level and model-level defenses, whose effectiveness depends on statistical properties of the model, deterministic enforcement layers use conventional, verifiable code to block prohibited actions regardless of what the LLM produces. Concrete examples include allowlists and blocklists for tool invocations, rate limits on sensitive operations such as financial transactions or file deletions, and regex or schema validation on tool arguments before execution.

\mypara{Summary}
The practices described above span a wide maturity spectrum. Input-level detection and model-level instruction hierarchy are active areas of academic research with early production deployments, but neither offers reliable standalone protection today. Deterministic enforcement mechanisms such as tool allowlists and human-in-the-loop confirmation are widely deployed in production agent systems and represent the most mature layer.
No single layer is sufficient on its own; the non-deterministic nature of LLM reasoning ensures that any individual defense can be circumvented under sufficiently adaptive attack strategies. The defense-in-depth composition of all three layers, where input-level defenses reduce attack volume, model-level defenses raise the bar for successful manipulation, and system-level defenses enforce hard limits on consequences, provides the most robust posture currently achievable.
We encourage NIST and CAISI to develop a layered-defense reference architecture for AI agent systems that practitioners can use as a checklist when designing, deploying, and auditing agent-based applications.

\section{Additional Considerations}\label{sec:3}

\begin{kkboxline}
\textbf{5(a):}
What methods, guidelines, resources, information, or tools would aid the AI ecosystem in the rapid adoption of security practices affecting AI agent systems and promoting the ecosystem of AI agent system security innovation?
\end{kkboxline}

Industry standards for agent communication, such as Model Context Protocol (MCP) and Agent2Agent Protocol (A2A), have begun to emerge. However, their security provisions primarily address low-level mechanisms such as authentication and transport security. They do not yet address higher-level security challenges in autonomous and multi-agent environments, including secure delegation, inter-agent trust boundaries, and privilege management across agents.

Security support in current agent development frameworks is still evolving and remains less mature than that of traditional software platforms. While many frameworks provide basic safeguards such as tool allowlists, sandboxed execution, and guardrail-based filtering, these frameworks generally lack comprehensive security models for privilege separation among agents, authorization for inter-agent interactions, and controls over delegation across chains of agents.

To promote the rapid adoption of secure practices, NIST and CAISI could develop guidance on security architectures for agent systems. Such guidance could include recommended security abstractions suitable for AI agents, best practices for secure multi-agent coordination, and reference models for privilege management, delegation control, and inter-agent trust relationships.

\begin{kkboxline}
\textbf{5(c):}
In which critical areas should research be focused to improve the current state of security practices affecting AI agent systems?
\end{kkboxline}

\mypara{Metrics and Benchmarks}
One important research direction is the development of metrics and benchmarks for evaluating the security of large language models and agentic systems. Security improvements are difficult to achieve without reliable methods for measuring them; what cannot be measured cannot be systematically improved. Establishing rigorous and widely accepted evaluation methodologies would enable researchers and developers to compare defenses, track progress, and identify remaining vulnerabilities.

Recent frontier models have shown improvements in their ability to resist indirect prompt injection attacks, and continued advances in this area remain an important component of securing AI agent systems. Strong and well-designed benchmarks can play a critical role in driving further improvements in LLM security. In addition to model-level robustness, other layers of defense would also benefit from improved metrics and benchmarks. These include mechanisms for detecting and filtering attacks, identifying malicious tools or agent capabilities, and evaluating system-level defenses designed to limit the impact of adversarial inputs.

Security benchmarks for agentic systems should emphasize dynamic interaction with realistic environments rather than static test suites, and should ideally incorporate adaptive adversaries. 
Static benchmarks can paint an overly optimistic picture of system security because defenses that block known attack patterns often fail when confronted with open-ended, multi-step attack trajectories. Robust security evaluation should therefore include adaptive adversarial testing, in which attack strategies evolve in response to the specific defenses being evaluated. Such dynamic and adversarial benchmarking would provide a more realistic assessment of the resilience of AI agent systems.

\mypara{Access Control Policies and Models}
Research should prioritize defense-in-depth architectures for AI agent systems, including at least one deterministic enforcement layer whose policy evaluation does not rely on LLM reasoning. 
The inherently non-deterministic nature of LLM-driven logic means that no single model-based safeguard can provide strong guarantees in isolation. Deterministic controls---such as sandbox boundaries, tool permissions, and action policies---can instead impose enforceable limits on agent behavior and complement probabilistic defenses.

An important research direction is identifying access control models suitable for specifying policies in this deterministic layer. Although fine-grained permissions may appear to improve security, they also require complex policies that are difficult to specify, verify, and maintain at scale.
As agents increasingly perform human-legible functions, established authorization frameworks for human access management provide a natural starting point. In particular, role-based access control (RBAC)~\cite{ferraiolo1992role,sandhu1996role,ferraiolo2001proposed}, originally developed by NIST researchers, offers a mature and scalable model for organizing permissions around functional roles.
However, RBAC alone may be insufficient for agent systems operating in dynamic environments. We believe a promising direction is to combine RBAC with risk-adaptive access control approaches~\cite{jason2004horizontal,cheng2007fuzzy,mcgraw2009radac}, in which authorization decisions are subject to limitations on aggregated risks. Hybrid approaches that integrate role-based structure with quantitative risk control may provide a practical foundation for deterministic authorization policies governing AI agents.

\mypara{Human Factors}
A common safeguard in current AI agent systems is to require human confirmation before executing actions deemed critical. While such mechanisms can reduce the likelihood of harmful actions, excessive confirmation requests can undermine usability and automation. Frequent prompts may lead to user fatigue, causing users to approve requests without careful review, thereby weakening the intended security protection. This tension highlights an important usable security challenge in agentic systems.

Research is needed to develop principled approaches for human-agent governance, determining how users should be involved in security-sensitive decisions while preserving the benefits of automation. One promising direction is risk-aware autonomy, in which users specify risk tolerance policies and the agent requests confirmation only when the estimated risk of an action exceeds a user-defined threshold. Another direction is enabling agent systems to learn user preferences and approval patterns from prior interactions, allowing them to better predict when explicit human oversight is necessary.

For agent systems, real-time confirmation may also be complemented by periodic transparency mechanisms, such as summaries or snapshots of actions taken, associated risks, and policy decisions. Such mechanisms can allow users to maintain situational awareness and oversight without being overwhelmed by frequent interruptions, thereby improving both security and usability in agentic environments.

\section{Conclusion}
Drawing on our operational and research experience, we believe that key elements of AI agent security include:
\begin{enumerate}
    \item Comprehensive threat modeling across model, tool, architecture, and hosting boundaries.
    \item Defense-in-depth architectures that incorporate at least one deterministic enforcement layer.
    \item Dynamic, adaptive evaluations that measure security outcomes over real multi-step workflows.
\end{enumerate}

Perplexity will continue to contribute both research insights and operational expertise to advance the security of AI agent systems. We look forward to continued collaboration with NIST/CAISI and our fellow stakeholders on these important issues.

\bibliography{reference}

\end{document}